\documentclass[10pt,twocolumn,letterpaper]{article}

\usepackage[pagenumbers]{cvpr} % To force page numbers, e.g. for an arXiv version

\usepackage[utf8]{inputenc} % 确保使用 UTF-8 编码
\usepackage[T1]{fontenc}    % 改善字体编码

\usepackage{url}
\usepackage{amssymb}
\usepackage{booktabs} % For toprule, midrule, and bottomrule
\usepackage{multirow} % For multi-row cells
\usepackage{multirow} % For multi-row cells
\usepackage{colortbl} % For cell coloring
\usepackage{xcolor}   % For custom colors
\usepackage{pifont}   % For special symbols (e.g., diamonds)
\usepackage{graphicx} % For including images
\usepackage{tabularx} % For automatic column width adjustment
\usepackage{pifont} % For uniform checkmarks and crosses
\newcommand{\cmark}{\ding{51}} % Checkmark symbol
\newcommand{\xmark}{\ding{55}} % Cross symbol
\usepackage{makecell}

\usepackage[accsupp]{axessibility}  % Improves PDF readability for those with disabilities.

\usepackage[most]{tcolorbox} 
\tcbset{
    colback=gray!20,     % 背景颜色
    arc=5mm,               % 圆角半径
    boxrule=0.3mm,         % 边框粗细
    width=\linewidth,      % 让盒子充满当前行宽
    left=2mm,              % 内左边距
    right=2mm,             % 内右边距
    top=2mm,               % 内上边距
    bottom=2mm,            % 内下边距
}

\usepackage{listings}
\usepackage{color}
\definecolor{codegreen}{rgb}{0,0.6,0}
\definecolor{codegray}{rgb}{0.5,0.5,0.5}
\definecolor{codepurple}{rgb}{0.58,0,0.82}
\definecolor{backcolour}{rgb}{0.95,0.95,0.92}
\lstdefinestyle{mystyle}{
    backgroundcolor=\color{backcolour},   
    commentstyle=\color{codegreen},
    keywordstyle=\color{magenta},
    numberstyle=\tiny\color{codegray},
    stringstyle=\color{codepurple},
    basicstyle=\ttfamily\scriptsize,
    breakatwhitespace=false,         
    breaklines=true,                 
    captionpos=b,                    
    keepspaces=true,                 
    numbers=left,                    
    numbersep=5pt,                  
    showspaces=false,                
    showstringspaces=false,
    showtabs=false,                  
    tabsize=2,
    showlines=true,
    morestring=[s]{'}{'},    % 单引号字符串
    morestring=[s]{"}{"},    % 双引号字符串
    literate={\'}{{\textquotesingle}}1  % 正确处理撇号
}

\newcommand\blfootnote[1]{% 
    \begingroup 
    \renewcommand\thefootnote{}\footnote{#1}% 
    \addtocounter{footnote}{-1}% 
    \endgroup 
}
\usepackage[hang,flushmargin]{footmisc}

\definecolor{cvprblue}{rgb}{0.21,0.49,0.74}
\usepackage[pagebackref,breaklinks,colorlinks,allcolors=cvprblue]{hyperref}

\def\paperID{****} % *** Enter the Paper ID here
\def\confName{CVPR}
\def\confYear{2026}

\title{Weaver: End-to-End Agentic System Training for Video Interleaved Reasoning}

\author{Yudi Shi$^{1,2*}$, Shangzhe Di$^{1}$, Qirui Chen$^{1}$, Qinian Wang$^{1}$, \\[2pt]
Jiayin Cai$^{2}$, Xiaolong Jiang$^{2}$, Yao Hu$^{2}$, Weidi Xie$^{1,\dagger}$ \\[3pt]
$^{1}$School of Artificial Intelligence, Shanghai Jiao Tong University, China\\[2pt]
$^{2}$Xiaohongshu Inc., China\\[2pt] \\[-12pt]
\textbf{\url{https://zhengrongz.github.io/Weaver/}}
}

\begin{document}
\maketitle
\blfootnote{
*: Work was done during internship in Xiaohongshu. \\
$\dagger$: Corresponding author.
}

\begin{figure*}[t]
  \centering
  \resizebox{.99\linewidth}{!}{
  \includegraphics{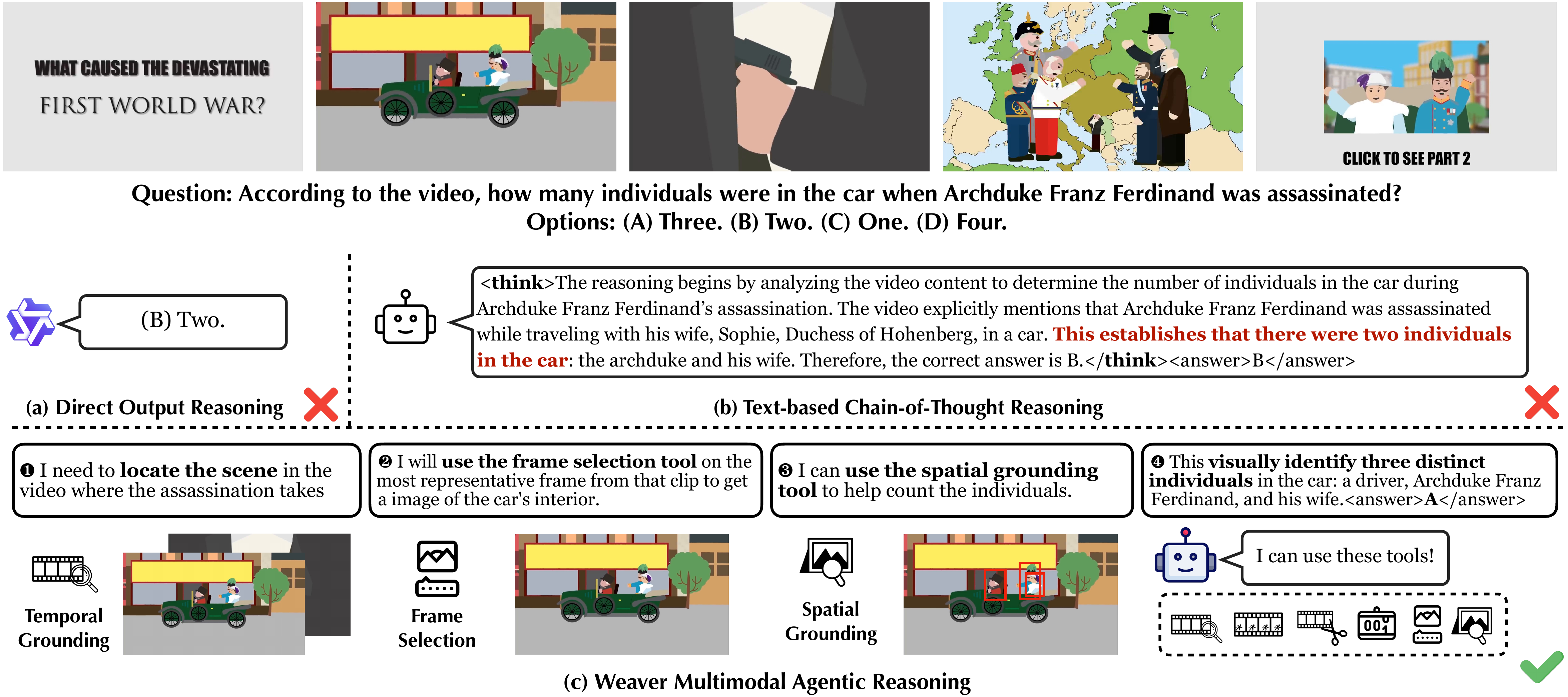}
  }
  \vspace{-5pt}
  \caption{
Our method, \textbf{Weaver}, leverages an interleaved visual-text reasoning paradigm, enabling the flexible combination and invocation of tools to progressively acquire visual information and generate multimodal reasoning trajectories towards final answer. As shown in (c), in comparison to the baseline methods illustrated in (a) and (b), Weaver successfully utilizes both the frame selection and spatial grounding tools to obtain a precise highlighted bounding box for the counting problem, which demonstrates the superiority of our approach.
  \vspace{-1.5em}
  }
  \label{fig: teaser}
\end{figure*}

\begin{abstract}
Video reasoning constitutes a comprehensive assessment of a model's capabilities, as it demands robust perceptual and interpretive skills, thereby serving as a means to explore the boundaries of model performance. While recent research has leveraged text-centric Chain-of-Thought reasoning to augment these capabilities, such approaches frequently suffer from representational mismatch and restricted by limited perceptual acuity. To address these limitations, we propose \textbf{Weaver}, a novel, end-to-end trainable multimodal reasoning agentic system. Weaver empowers its policy model to dynamically invoke diverse tools throughout the reasoning process, enabling progressive acquisition of crucial visual cues and construction of authentic multimodal reasoning trajectories. Furthermore, we integrate a reinforcement learning algorithm to allow the system to freely explore strategies for employing and combining these tools with trajectory-free data. Extensive experiments demonstrate that our system, Weaver, enhances performance on several complex video reasoning benchmarks, particularly those involving long videos.
\end{abstract}    
\vspace{-1.5em}
\section{Introduction}
\label{sec:intro}
Video reasoning~\cite{cheng2025videoholmes,zhang2025tinyllava,cheng2025vstar,wang2025videorts,park2025deepvideo, dang2025reinforcing, li2025videochatr1} plays a crucial role in the pursuit of Artificial General Intelligence~(AGI). It requires models to extract reliable information from complex video scenes and to perform multi-step spatio-temporal inference that culminates in correct answers. Unlike verbal or image-level reasoning, video understanding presents a multifaceted challenge, involving not only straightforward logical analysis but also accurate spatio-temporal perception. This process further depends on foundational capabilities like temporal grounding and spatial tracking~\cite{unitime2025, liu2024grounding}.

Recent efforts~\cite{shi2025enhancing,han2025videoespresso,fei2024vot} in video understanding have focused on fine-tuning multimodal large language models (MLLMs) using step-wise rationales—either synthetic or human-authored—to enhance temporal reasoning~\cite{li2025llavast}, event decomposition~\cite{shi2025enhancing}, and long-form QA~\cite{zohar2024videostar}. Despite these advances, existing video Chain-of-Thought pipelines~\cite{feng2025videor1,wang2025videorft,chen2025longvila-r1, chen2025versavid} remain text-centric: intermediate states are represented as text and computed without adaptive access to visual evidence. This leads to three persistent limitations: (i) hallucination: as textual rationales lengthen, the model drifts from the video signal, yielding fabricated objects, actions, or attributes;
(ii) frozen perception: models typically reason over a fixed subset of sampled frames or pre-extracted features, lacking mechanisms to iteratively acquire additional information during reasoning;
(iii) representational mismatch: textual rationales cannot faithfully encode structured visual signals ({\em e.g.}, segmentation masks, depth, optical flow, tracks), limiting access to geometric and pixel-level evidence crucial for reliable conclusions. These issues motivate a shift from text-only rationales toward agentic, perception-in-the-loop multimodal reasoning trajectories that interleave reasoning with visual tool use. 

In this paper, we propose to develop an end-to-end multimodal reasoning agentic system that distinguishes itself through flexible and unrestricted exploration of tool utilization. Our approach augments the core reasoning model with a curated library of specialized perception tools, each functioning as an expert for a specific subtask ({\em e.g.}, detection, tracking, temporal localization). During inference, the core model engages in multi-turn, interleaved visual-text reasoning; it analyzes the question, selectively invokes appropriate tools to obtain targeted visual evidence, and subsequently integrates the retrieved signals into ongoing process. The final reasoning trajectory is embodied in the form of a multi-modal, visual-text interleaved Chain-of-Thought. This agentic, perception-in-the-loop paradigm facilitates progressive evidence acquisition, mitigates hallucination, and ensures that intermediate representations are aligned with the true structure of visual information.

To instill this capability, we adopt a two-stage training strategy: 
(i) we conduct cold-start supervised finetuning (SFT) to teach basic tool invocation, argument formatting, and interleaved reasoning patterns; 
(ii) we then perform end-to-end, tool-augmented reinforcement learning (RL), allowing the model to explore tool compositions and trajectories that optimize task performance under realistic decision-making dynamics. To support these stages, we introduce two datasets tailored for interleaved reasoning: \textbf{\texttt{Weaver-SFT-10K}} for supervised instruction of tool usage and \textbf{\texttt{Weaver-RL-12K}} for reinforcement learning with tool-augmented reward.

Our proposed multimodal agentic system, termed as \textbf{Weaver}, achieves consistent gains on diverse video understanding benchmarks spanning long-form reasoning, perception, and spatial tasks. Notably, Weaver surpasses the base model by 6.7\%, 2.4\%, and 2.6\% on LVReason, LVBench, and MLVU, respectively, highlighting the benefits of progressive, tool-mediated evidence acquisition in long videos. Further ablation studies show that, 
our system has effectively acquired the capability to combine tools for problem-solving across various task types. Moreover, the model can be efficiently trained at scale via reinforcement learning using trajectory-free data, highlighting the potential for scalable training.

In summary, our contributions are threefold: 
(i) we propose a multimodal agentic system that enables interleaved reasoning for complex video tasks, that is trained end-to-end with reinforcement learning; 
(ii) We construct two high-quality reasoning datasets for training: \textbf{\texttt{Weaver-SFT-10K}} and \textbf{\texttt{Weaver-RL-12K}}; 
(iii) Through extensive experiments, we demonstrate that our agentic system consistently outperforms existing approaches on a range of long video understanding and general video question answering benchmarks.
\begin{figure*}[t]
  \centering
  \resizebox{\linewidth}{!}{
  \includegraphics{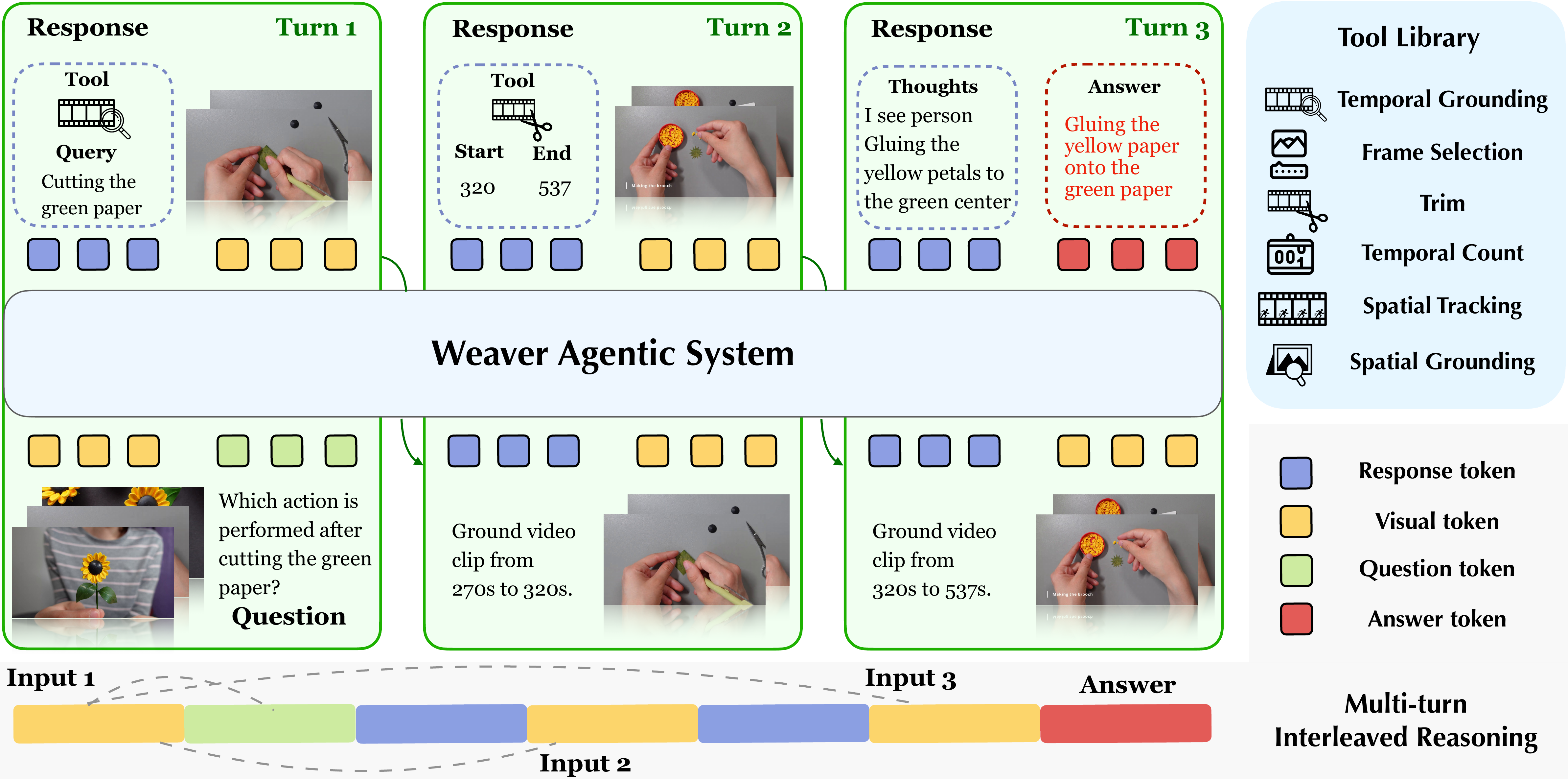}
  }
  % \vspace{-15pt}
  \caption{
    Overview of \textbf{Weaver} agentic system. During the multi-turn interleaved reasoning process, Weaver concatenates all tokens generated in previous rounds as input for subsequent rounds, continuing this procedure until a final answer is obtained. Consequently, the entire reasoning process can be interpreted as a multi-round conversational exchange.
  }  
  \label{fig: architecture}
  \vspace{-1.5em}
\end{figure*}

\section{Related Work}
\label{sec: related}

\vspace{2pt}\noindent\textbf{Multimodal Large Language Model.}
Recent Multimodal Large Language Models (MLLMs) have demonstrated remarkable capabilities in video perception and understanding. Both proprietary models, such as Gemini-2.5-Pro~\cite{comanici2025gemini}, and open-source models, including Qwen2.5-VL~\cite{bai2025qwen2_5vl} and InternVL-3.5~\cite{wang2025internvl3}, are able to process sequences composed of hundreds or even thousands of frames, thereby enabling complex tasks such as video captioning and question answering.
However, the prevailing strategy of uniformly sampling frames and directly generating answers lacks advantages in both resource efficiency and accuracy when applied to intricate video reasoning scenarios. This limitation has motivated researchers to explore new approaches to further enhance the performance of these models.

\vspace{2pt}\noindent\textbf{Agentic Reasoning Methods.}
With advances in large language models (LLMs) and various computer vision subfields, recent works~\cite{gupta2023visual,suris2023vipergpt,fan2024videoagent} have begun constructing agentic systems that leverage LLMs as planners, while employing expert models as tools to solve complex questions. These approaches typically decompose a question into explicit sub-tasks and then invoke specialized tools to address them; the LLM subsequently summarizes the final answer based on the step-wise results.

These agentic methods demonstrate how powerful language models can be integrated with vision experts. However, these systems are generally training-free, relying on in-context examples to guide the generation process, and lack the ability for autonomous exploration and adaptation.

\vspace{2pt}\noindent\textbf{Multimodal Chain-of-Thought Reasoning.}
Chain of thought (CoT) refers to the explicit output of intermediate reasoning steps when solving problems, a technique that has been shown to improve models' performance on complex tasks. In the context of visual understanding, many studies~\cite{shao2024visualcot,fei2024vot, liu2025videomind,han2025videoespresso, huang2025visionr1, bai2025univg, liu2025seg} have adopted CoT to enhance model interpretability and problem-solving abilities. However, obtaining high-quality trajectory annotations is often expensive and labor-intensive, prompting a shift towards reinforcement learning-based training paradigms.

\vspace{2pt}\noindent\textbf{Concurrent Work.}
Recently, the rise of ``think with images'' approaches~\cite{openai2025gpto3} has sparked a transition from purely text-based to multimodal CoTs, in which visual information is incorporated into the reasoning process to further support model inference, which is similar to our multimodal reasoning trajectory. Inspired by this approach, several studies have attempted to reproduce similar methodologies within the domains of image~\cite{zhou2025revpt, zheng2025deepeyes,fan2025grit,zhang2025cof} and video~\cite{xie2025videomtr,zhang2025vital} reasoning. 
For instance, ViTCoT~\cite{zhang2025vitcot} leverages a pre-extracted interleaved format of reasoning, demonstrating the efficacy of multimodal CoT. 
Methods such as VITAL~\cite{zhang2025vital} and Video-MTR~\cite{xie2025videomtr} implement interleaved reasoning by inserting temporally-grounded video clips into model conversations as auxiliary visual context. 
However, existing approaches mostly rely on one single model or perception tool, which restricts their problem-solving capabilities. 
In contrast, our proposed multimodal agentic system can expand comprehensive tool library to solve problems.

\begin{table*}[t]
\caption{Visual tools used in Weaver. We select recent state-of-the-art models for each tool. These models are kept frozen throughout the entire training process to preserve their capabilities.
}
\label{tab:visual tools}
\vspace{-5pt}
\centering
\renewcommand{\arraystretch}{1.2} % Adjust row height
\setlength{\tabcolsep}{1.1mm}      % 控制列间距
\footnotesize
% \small
% \vspace{1em}
% \resizebox{\linewidth}{!}{
\begin{tabular}{lclll}
\toprule
Tool & Model & Description & Input & Output \\
\midrule
Temporal Grounding~($\Phi_{\text{TG}}$) & UniTime~\cite{unitime2025} & Ground video clip temporally according to query & Video clip, Query & Grounded video clip \\
Frame Selection~($\Phi_{\text{FS}}$) & Qwen2.5-VL~\cite{bai2025qwen2_5vl} & Select most representative frame according to query & Video clip, Query & Representative frame \\
Temporal Count~($\Phi_{\text{TC}}$) & Qwen2.5-VL~\cite{bai2025qwen2_5vl} & Judge and merge video clips where query occurs & Video clip, Query & Spliced video clips \\
Trim~($\Phi_{\text{TR}}$) & / & Ground video clip according to start and end & Video clip, Start, End & Grounded video clip \\
Spatial Tracking~($\Phi_{\text{ST}}$) & GroundedSAM2~\cite{ren2024grounded} & Track target objects in video & Video clip, Objects & Highlighted video clip \\
Spatial Grounding~($\Phi_{\text{SG}}$) & Grounding-Dino~\cite{liu2024grounding} & Ground objects spatially per frame & Video clip, Objects & Highlighted video clip \\
\bottomrule
\end{tabular}
\vspace{-1.5em}
\end{table*}

\section{Method}
\label{sec: method}

This section details our proposed \textbf{Weaver}, a multimodal agentic system that equips video large language models (Video-LLMs) with interleaved, perception-in-the-loop reasoning.
Specifically, \textbf{Weaver} endows a core reasoning model with a library of specialized visual perception tools, and is trained with reinforcement learning to enable dynamic tool calling and interleaved visual-text reasoning. At inference time, the core model analyzes the question , selectively invokes tools to acquire targeted intermediate visual evidence, {\em e.g.}, detections, tracks, temporally grounded clips, integrates the returned signals back into its ongoing chains, and finally produces a multimodal reasoning trajectory that leads to the final answer.

\subsection{Problem Formulation}
We formulate our multimodal agentic system under a dynamic decision-making framework, 
that enables step-wise thinking and tool calling. 
The whole system consists of the following components: 

\begin{itemize}
  \setlength\itemsep{0.3em}
  \item \textbf{Policy model~($\mathcal{M_\theta}$)}: 
  A trainable generative vision-language model~(VLM) for stepwise reasoning, 
  {\em e.g.}, Qwen2.5-VL.
  \item \textbf{Tool library~($\mathcal{T}$)}: 
  A set of tools for conducting visual perception tasks, as detailed in Table~\ref{tab:visual tools}, we select six tools for both dynamic and static scenes. The primary criterion for selecting tools is their ability to perceive spatio-temporal information, which is essential for addressing complex video reasoning tasks,
  {\em i.e.}, $\mathcal{T} = \{\Phi_{\text{TG}}(\cdot),\Phi_{\text{FS}}(\cdot),\Phi_{\text{TR}}(\cdot),\Phi_{\text{TC}}(\cdot),\Phi_{\text{ST}}(\cdot),\Phi_{\text{SG}}(\cdot)\}$.
  \item \textbf{History state~($\mathcal{H}$)}: A set that comprises all history textual response, and visual information until certain step, 
  {\em i.e.}, $\mathcal{H}_i = \{s_0, s_1,\dots, s_i\}$.
\end{itemize}

In practise, given a raw video $\mathcal{V} \in \mathbb{R}^{T \times H \times W \times 3}$ and a question $\mathcal{Q}$, our objective is to train the policy model $\mathcal{M_\theta}$, 
that enables to dynamically invoke tools from $\mathcal{T}$, 
and progressively reasoning the final answer with the integrated textual responses and visual information.

\vspace{3pt} \noindent \textbf{Initial state~($\mathcal{H}_0$).}
We first define $\mathcal{H}_0$ as follows:
\begin{equation}
    \mathcal{H}_0 = \{s_0\} = (\mathcal{Q}, v_0)
\end{equation}
where $v_0$ is a uniformly sampled clip extracted from $\mathcal{V}$.

\vspace{3pt} \noindent \textbf{Stepwise reasoning~($s_i$).}
In order to generate new step, the model incorporates all preceding steps as historical states and produces the next step iteratively, until the final answer is obtained:
\begin{equation}
\begin{aligned}
    s_i &= \mathcal{M_\theta}(\mathcal{H}_{i-1}, \mathcal{V}, \mathcal{T}) \\
        &= \mathcal{M_\theta}(\mathcal{Q},v_0,\dots,r_{i-1}, v_{i-1}, \mathcal{V}, \mathcal{T})
\end{aligned}
\label{eq:gene}
\end{equation}
where each step results with a textual response $r_i$, and potentially an intermediate video clip $v_i$:

\begin{equation}
s_i = 
\begin{cases}
(r_i, v_i), & \text{if tool\_call in } r_i \\
(r_i, \emptyset), & \text{otherwise}
\end{cases}
\end{equation}
Here, $v_i \in \mathbb{R}^{t_i \times H \times W \times 3}$ is obtained through tool invocation, as described in Sec.~\ref{sec:tool-calling}.
Once a new step is generated, the model employs the structured format template \texttt{<answer>} \dots \texttt{</answer>} to identify the final answer, which may consist of either predefined options or open-ended responses, depending on the specific requirements of the task.
If the newly generated step $s_i$ does not yield the final answer, $\mathcal{M_\theta}$ proceeds to the next round of generation. 

\vspace{3pt} \noindent \textbf{State update~($\mathcal{H}_{i-1} \rightarrow \mathcal{H}_i$).}
In this case, the new step $s_i$ is incorporated into $\mathcal{H}_{i-1}$, forming the updated state $\mathcal{H}_i$, and the generation process is repeated. All previous steps are utilized as context, allowing the model to refer to the entire interaction history during subsequent reasoning.

The following sections will detail the stepwise reasoning process. 
Sec.~\ref{sec:encoding} presents how we convert the textual responses and visual observations in $\mathcal{H}_{i-1}$ into an interleaved input sequence suitable for processing by $\mathcal{M}_\theta$, yielding the next textual response $r_i$. Subsequently, Sec.~\ref{sec:tool-calling} details how new visual information $v_i$ is acquired via tool invocation.

\begin{figure*}[t]
  \resizebox{\linewidth}{!}{
  \includegraphics{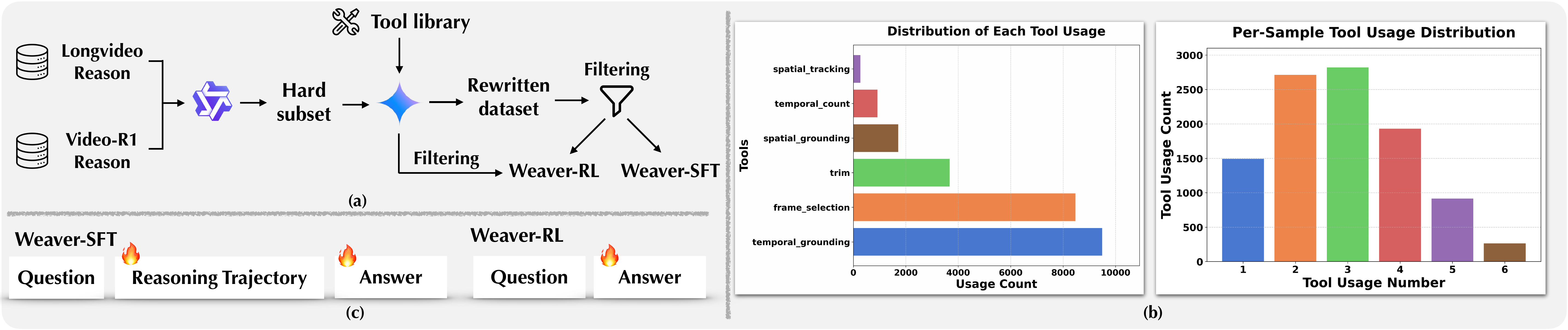}
  }
  % \vspace{-15pt}
  \caption{
    Data Pipeline and Statistics. Panel (a) illustrates the data construction pipeline for \textbf{\texttt{Weaver-SFT}} and \textbf{\texttt{Weaver-RL}}, beginning with two textual CoT reasoning datasets and resulting in two high-quality reasoning datasets. Panel (b) presents the statistical analysis of Weaver-SFT. Panel (c) shows the specific compositions of two dataset, fire emojis mean the content will be supervised during training.
    }
  \label{fig: data pipe}
  \vspace{-1.5em}
\end{figure*}

\subsection{Vision-Language Interleaved Encoding}\label{sec:encoding}
Assume that we have already accumulated the history state $\mathcal{H}_{i-1}=\{\mathcal{Q},v_0,\dots,r_{i-1},v_{i-1}\}$, that consists of video clips in each step: $\{v_0, v_1,\dots, v_{i-1}\}$, and textual responses of each step: $\{\mathcal{Q}, r_1, \dots, r_{i-1}\}$. These sequences are then processed separately: the video sequence is fed into the vision encoder $\Phi_{\text{vis}}$ of $\mathcal{M_\theta}$, while the response sequence is input to the language model $\Phi_{\text{text}}$ of $\mathcal{M_\theta}$ to obtain the visual features and token sequences respectively:
\begin{equation}
\begin{aligned}
&\{\mathcal{F}_0, \mathcal{F}_1, \dots, \mathcal{F}_{i-1}\} = \Phi_\mathrm{vis}(\{v_0, v_1, \dots, v_{i-1}\}), \\
&\{\sigma_0, \sigma_1, \dots, \sigma_{i-1}\} = \Phi_\mathrm{text}(\{\mathcal{Q}, r_1, \dots, r_{i-1}\})
\end{aligned}
\label{eq: encode}
\end{equation}
where $\mathcal{F}_i \in \mathbb{R}^{n_i \times D}$ denotes the visual feature of video clip $v_i$, and $\sigma_i \in \mathbb{R}^{m_i \times D}$ represents the token sequence corresponding to text response $r_i$. Note that, in implementation, we can actually cache the corresponding results at each step, so only the content generated in the most recent round needs to be newly encoded in subsequent rounds.

As illustrated in Figure~\ref{fig: architecture}, the model sequentially inserts the visual features and token sequences, maintaining their original order to construct the final interleaved input sequence $\mathcal{I}_{i-1} \in \mathbb{R}^{N \times D}$:
% \vspace{-5pt}
\begin{equation}
    \mathcal{I}_{i-1} = \{\sigma_0, \mathcal{F}_0, \sigma_1, \dots, \sigma_{i-1}, \mathcal{F}_{i-1}\}
\end{equation}
$N$ refers to the total number of visual and textual tokens.

Noted that $\mathcal{I}_{i-1}$ is encoded from $\mathcal{H}_{i-1}$ and is regarded as  the processed interleaved input to $\mathcal{M_\theta}$ for generating new textual response $r_i$:

\begin{equation}
    r_i = \mathcal{M_\theta}(\mathcal{I}_{i-1})
\end{equation}
The generated response $r_i$ is then employed for tool invocation and becomes a part of the new step $s_i$.

\subsection{Tool Calling}\label{sec:tool-calling}

After producing a textual response $r_i$, the model inspects the output for potential tool invocations using a structured template of the form \texttt{<tool\_call>} \dots \texttt{</tool\_call>}. Any content enclosed within this template must specify (i) the tool name and (ii) its input arguments, thereby fully determining how the tool should be executed to retrieve the intermediate video clip $v_i$.

\begin{equation}
    v_i = \mathcal{T}(\mathcal{V},\mathcal{N}, \text{argument}) \in \mathcal{V}
\end{equation}
where $\mathcal{N}$ denotes the name of the tool to invoke and arguments specifies its input parameters.
If no tool call is extracted from the response, the policy model $\mathcal{M_\theta}$ is deemed to have chosen not to invoke any tool at this step.

The resulting step $s_i=\{r_i, v_i\}$ can then be consumed either to produce the final answer or to generate the next step in the trajectory.

\subsection{Model Training}
To facilitate the core model's ability to flexibly utilize and combine tools for interleaved visual-text reasoning, we employ a two-stage training regime. 
The first stage involves supervised finetuning for cold start, while the second stage leverages tool-augmented reinforcement learning with trajectory-free data. 
Accordingly, we have constructed two training datasets: 
\textbf{\texttt{Weaver-SFT-10K}} and \textbf{\texttt{Weaver-RL-12K}}.

\begin{table*}[t]
\caption{Main experiment results on various video benchmarks. Weaver achieves superior performances compared to other models especially in long video benchmarks. * means results reproduced by ourselves and other results are retrieved from original papers.
}
\label{tab:mc performance}
\vspace{-5pt}
\centering
\renewcommand{\arraystretch}{1.1} % Adjust row height
\setlength{\tabcolsep}{2.5mm}      % 控制列间距
\footnotesize
% \small
% \vspace{1em}
% \resizebox{\linewidth}{!}{
\begin{tabular}{lcccccccc}
\toprule
\textbf{Model} & \multirow{2}{*}{\textbf{\#Frames}} & \textbf{LVReason} & \textbf{VideoMME} & \textbf{LVBench} & \textbf{MLVU}& \textbf{VideoMMMU}  & \textbf{VSIBench} & \textbf{MVBench}\\
Duration(s) &  &  424 & 1010 & 4101 & 934 & 507 & 97 & 16\\
\midrule
\multicolumn{9}{l}{\textbf{Proprietary Models}} \\
Gemini-1.5-Pro \cite{team2024gemini1.5} & 1fps & 67.3 & 75.0 & 33.1 & - & 53.9 & 45.4 & 60.5\\
GPT4-V \cite{openai2023gpt4v} & 1fps & - & 59.9 & - & - & - & - & 43.7\\
GPT4-o \cite{openai2024gpt4o} & 1fps & 60.7 & 71.9 & 30.8 & 54.9 & 61.2 & 34.0 & 64.6\\
\midrule
\multicolumn{9}{l}{\textbf{Open-source Direct Base Models}} \\
LLaVA-OneVision~\cite{li2024llavaov} & 32 & - & 58.2 & - & - & 34.4 & 32.4 & 56.7 \\
InternVL2.5~\cite{chen2024internvl2.5} & 16-64 & - & 64.2 & 38.4 & - & - & - & 72.0 \\
Qwen2.5-VL~\cite{bai2025qwen2_5vl} & 128 & 68.7\rlap{\textsuperscript{*}} & 63.3\rlap{\textsuperscript{*}} & 40.6\rlap{\textsuperscript{*}} & 52.9\rlap{\textsuperscript{*}} & 46.6\rlap{\textsuperscript{*}} & 38.6\rlap{\textsuperscript{*}} & 66.0\rlap{\textsuperscript{*}}\\ 
\midrule
\multicolumn{9}{l}{\textbf{Open-source Text-centric CoT Models}} \\
Video-R1~\cite{feng2025videor1} & 64 & 71.4\rlap{\textsuperscript{*}} & 61.4 & 40.5\rlap{\textsuperscript{*}} & 42.5\rlap{\textsuperscript{*}} & 52.4 & 37.1 & 64.8\\
VideoRFT~\cite{wang2025videorft} & 32 & 72.0\rlap{\textsuperscript{*}} & 59.8 & 41.1\rlap{\textsuperscript{*}} & 45.0\rlap{\textsuperscript{*}} & 51.1 & 36.8 & 62.1\\
Long-VILA-R1~\cite{chen2025longrl} & 512 & 67.9 & 65.1 & - & - & - & - & 67.6\\
\midrule
\multicolumn{9}{l}{\textbf{Open-source Tool-invoked Interleave Models}} \\
PixelReasonser~\cite{su2025pixel} & 16 & 71.4\rlap{\textsuperscript{*}} & - & 39.2\rlap{\textsuperscript{*}} & 43.9\rlap{\textsuperscript{*}} & - & - & 67.8\\
Video-MTR~\cite{xie2025videomtr} & 32 & - & 59.0 & - & 48.4 & - & - & -\\
FrameMind~\cite{ge2025framemind} & 32 & - & 60.9 & - & 48.6 & - & - & 64.2\\
\hline
\rowcolor{green!10}
\rowcolor{green!10} Weaver-SFT & 128+1fps & 71.7 & 63.8 & 40.1 & 51.7 & 48.8 & 38.0 & 66.5\\
\rowcolor{green!10} Weaver & 128+1fps & 75.4 & 65.3 & 43.0 & 54.5 & 51.3 & 40.3 & 67.7\\
\bottomrule
\end{tabular}
% }
\vspace{-1.5em}
\end{table*}

\subsubsection{Dataset Construction}

We curate training data from two high-quality video reasoning dataset with \textbf{textual Chain-of-Thought} annotations: Video-R1-170K~\cite{feng2025videor1} and LongVideo-Reason-51K~\cite{chen2025longrl}. The goal here is therefore to convert these textual ones into \textbf{tool-augmented multimodal reasoning trajectories.}

\vspace{2pt} \noindent \textbf{Preliminary filtering.}
As the first-pass filter, we use Qwen2.5-VL-7B to directly answer all questions and discard examples that are  correctly answered, thereby focusing on cases that can potentially benefit from step-wise reasoning and tool use.

\vspace{2pt} \noindent \textbf{Trajectory generation.}
For the remaining items, we prompt Gemini-2.5-Pro~\cite{comanici2025gemini} with (i) detailed documentation of our tool library and (ii) the original textual CoT, instructing it to rewrite the reasoning trajectory to explicitly incorporate tool invocations. We then pre-extract the intermediate visual information referenced in these rewritten trajectories using our tool library.

\vspace{2pt} \noindent \textbf{Final refinement.}
Here, we supply these rewritten trajectory—excluding the final ground-truth answer—as context to Qwen2.5-VL,
which is tasked to produce the final answer solely from this provided context. 
We retain instances that model answers correctly as the supervised fine-tuning (SFT) dataset, thereby filtering residual failure cases. The remaining subset is collected as the seed pool for downstream reinforcement learning (RL) training which only requires question-answer pairs and operates in a trajectory-free manner. The full pipeline is illustrated in Figure~\ref{fig: data pipe}.

\vspace{2pt} \noindent \textbf{Dataset statistics.}
Applying the above pipeline yields two high-quality datasets: (i) a tool-augmented reasoning trajectory set for cold-start supervised training, \textbf{\texttt{Weaver-SFT-10K}}, 
and (ii) a trajectory-free set for reinforcement learning, \textbf{\texttt{Weaver-RL-12K}}.

As shown in Figure~\ref{fig: data pipe}, we profile Weaver-SFT-10K along two axes: the overall frequency of tool usage across dataset and the per-sample distribution of tool invocations. Two patterns emerge from this analysis: (i) temporal tools dominate usage, underscoring the centrality of precise temporal grounding in video reasoning; (ii) the relatively high average number of tool calls per sample indicates that complex queries are typically solved via compositions of multiple tools.

\subsubsection{Cold Start}
We initialize the system with supervised fine-tuning (SFT) to bootstrap reliable tool use. Given the breadth of the tool library—spanning multiple tool families and nontrivial compositions—the base model benefits from explicit demonstrations of well-structured reasoning trajectories that coordinate tool calls. During this cold-start phase, token-level supervision enforces the target response format and instills core patterns for selecting, composing, and invoking tools across diverse problem settings.

Given a question $\mathcal{Q}$ and video $\mathcal{V}$, our training objective is to minimize the cross-entropy loss for policy model $\mathcal{M}_\theta$: 
\begin{equation}
\mathcal{L}(\theta) = -\frac{1}{N} \sum_{i=1}^{N} \sum_{t=1}^{n^i} \log P_\theta \left( s_t^i \mid \mathcal{H}_{t-1}^i, \mathcal{V}^i, \mathcal{T} \right)
\end{equation}
where $N$ is the training batch size, $n^i$ is the total steps of reasoning process.

\subsubsection{Reinforcement Learning}

Reinforcement learning (RL) supervises policies via outcome-level rewards rather than stepwise labels, substantially reducing annotation cost and avoiding compounding supervision errors along generated trajectories. This enables effective training on large-scale, trajectory-free corpora that contain only question–answer pairs. In addition, RL promotes exploration over tool choices and invocation strategies through stochastic sampling, improving the model’s ability to generalize across diverse problem types.

Following the cold-start stage, we adopt a tool-augmented variant of GRPO~\cite{guo2025deepseek}. Our modification augments the vanilla objective with an auxiliary reward that explicitly incentivizes successful tool use, thereby encouraging the model to integrate tools into its reasoning process when beneficial. The composite reward comprises three components: (i) correctness (agreement with the target answer), (ii) format (adherence to the required output specification), and (iii) tool usage (successful and appropriate invocation of tools).

\vspace{2pt} \noindent \textbf{Correctness.} 
To ensure the final answer is verifiable, we utilize multiple-choice questions for training. The reward $R_{\text{corr}}$ is evaluated according to the correctness of the output answer:
$R_{\text{corr}}=1$ if answer is correct, otherwise $R_{\text{corr}}=0$.

\vspace{2pt} \noindent \textbf{Format.} This reward ensures the model generates structured responses that are easily parsable for evaluation. A reward $R_{\text{format}}$ of 1 is given if the response contains the structured tag \texttt{<answer>} \dots \texttt{</answer>}, and 0 otherwise.

\vspace{2pt} \noindent \textbf{Tool usage.} Inspired by Deepeyes~\cite{zheng2025deepeyes}, we introduce a tool-successful-usage reward $R_{\text{tool}}$. This reward is designed to incentivize not just correctness, but the process of using tools to achieve it. Specifically, $R_{\text{tool}}$ is considered as 1 when the reasoning trajectory contains the \texttt{<tool\_call>} \dots \texttt{</tool\_call>} tag and the final answer is correct.

\vspace{2pt} \noindent \textbf{Total reward.} 
The total reward can therefore be computed as: $R_{\text{final}}=\lambda_1*R_{\text{corr}}+\lambda_2*R_{\text{format}}+\lambda_3*R_{\text{tool}}$. Here we set $\lambda_1=0.7$, $\lambda_2=0.2$, $\lambda_3=0.1$ to represent the weight of each reward. This naive but effective reward design helps model learn how to use tools correctly to solve problems.

\section{Experiments}
\label{sec: experiments}

\begin{figure*}[htbp] % 使用 figure* 环境横跨双栏
    \centering
    % 第一张子图
    % % 第二张子图
    \begin{subfigure}[b]{0.4\textwidth}
        \includegraphics[width=\textwidth]{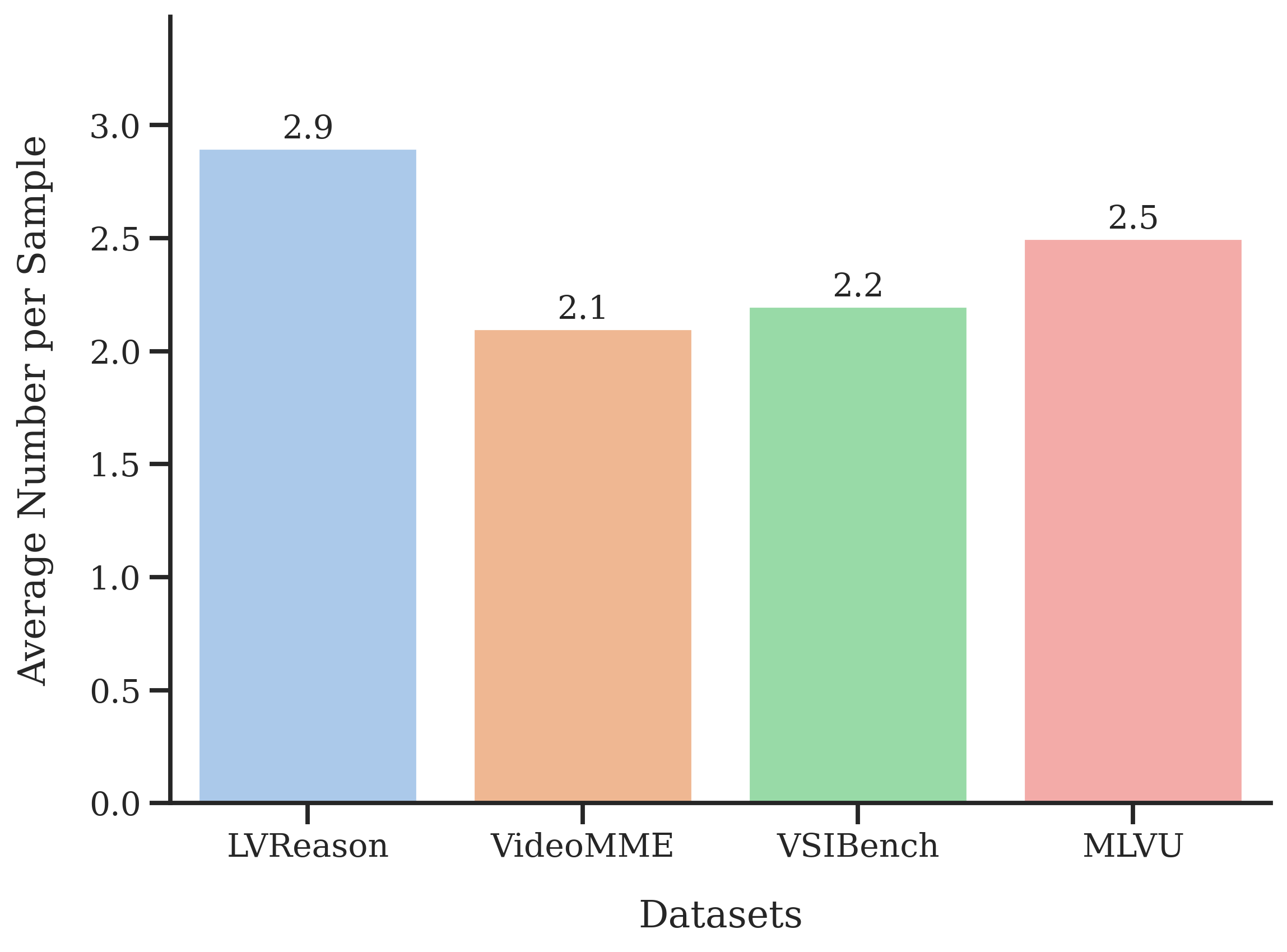}
        % \vspace{-5pt}
        \caption{Average Tool usage number across different benchmarks.}
        \label{fig:sub1}
    \end{subfigure}
    \hfill
    % 第三张子图
    \begin{subfigure}[b]{0.58\textwidth}
        \includegraphics[width=\textwidth]{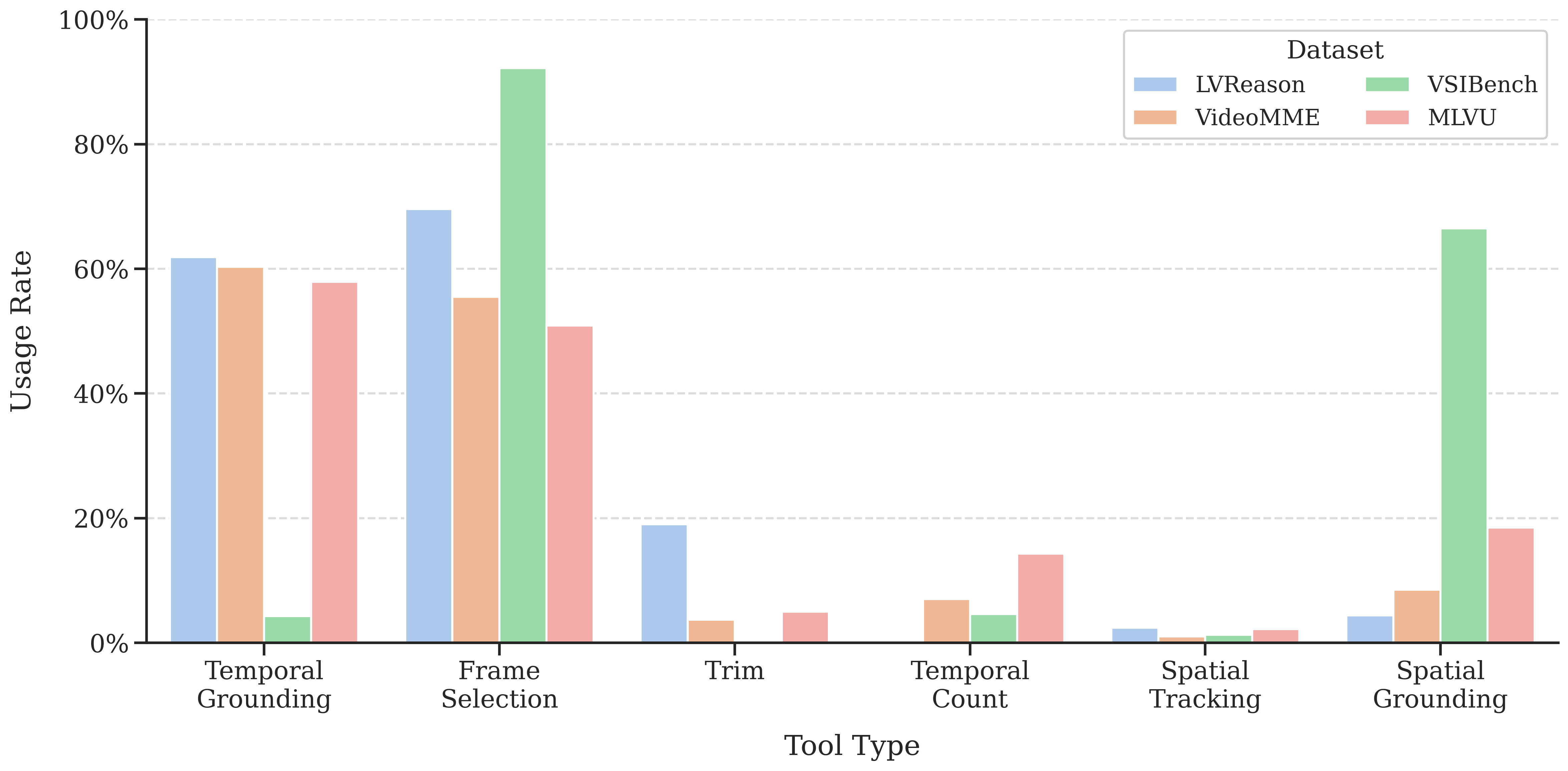}
        % \vspace{-5pt}
        \caption{Distribution of Different Tools usage across different benchmarks.}
        \label{fig:sub2}
    \end{subfigure}
    \vspace{-5pt}
    \caption{Tool Usage analysis for Weaver in different evaluation benchmarks.}
    \label{fig:eval tool analysis}
    \vspace{-1.5em}
\end{figure*}

\subsection{Setup}
\vspace{2pt} \noindent \textbf{Implementation details.}
We employ Qwen2.5-VL-7B~\cite{bai2025qwen2_5vl} as the core reasoning model due to its robust visual perception capabilities. During the cold-start stage, 64 frames are uniformly sampled from each video. For the RL training stage, the sampling strategy is adapted: we first uniformly sample 128 frames, and when tools are invoked, a subsequent sampling is performed at 1 FPS.

\vspace{2pt} \noindent \textbf{Evaluation benchmarks.}
We conduct extensive evaluations on various benchmarks. For video perception, we choose MVBench~\cite{li2024mvbench} for its rich variety of tasks and the range of exploration abilities. For Long video reasoning, we select LongVideo-Reason~\cite{chen2025longvila-r1}, MLVU~\cite{zhou2025mlvu} and LVBench~\cite{wang2025lvbench}. For general VideoQA, VideoMME~\cite{fu2025videomme} is a good choice for its complete setting. In addition, we also evaluate VideoMMMU~\cite{hu2025videommmu} and VSIBench~\cite{yang2025thinking} for their focuses on Knowlegde and spatial reasoning. It is noted that for MLVU, we report the results on test set and for VideoMME, we present the results without subtitles.

\subsection{Main Results}
As shown in Table~\ref{tab:mc performance}, we divide our analysis into three parts.

\vspace{2pt} \noindent \textbf{Comparison with base model.}
Compared to the base model Qwen2.5-VL, Weaver demonstrates superior performance across all benchmarks, achieving improvements of 6.7\% on LVReason, 4.7\% on VideoMMMU. The overall improvements indicate that our method effectively enhances the base model's reasoning ability. The SFT model does not show performance improvements on all benchmarks, which may be due to the implementation of strict token supervision, causing the system to be unable to solve new problem types it has not encountered. 

\vspace{2pt} \noindent \textbf{Comparison with text-centric CoT methods.}
Weaver also outperforms existing text-centric Chain-of-Thought methods on nearly all benchmarks, further validating the effectiveness of our interleaved reasoning agentic system. This advantage is particularly evident in long video benchmarks; Weaver achieves a 12\% improvement on MLVU compared to Video-R1 and a 9.5\% improvement compared to VideoRFT. This can be attributed to Weaver’s ability to progressively obtain new visual information.

\vspace{2pt} \noindent \textbf{Comparison with interleaved reasoning methods.}
Compared to other methods that employ interleaved reasoning, Weaver also achieves remarkable results across benchmarks. For example, Weaver demonstrates average improvements of 5.4\% on VideoMME and 5.0\% on MLVU over Video-MTR and FrameMind. These results suggest that the collaborative integration of multiple tools can further improve system effectiveness.

\begin{figure*}[t]
  \resizebox{\linewidth}{!}{
  \includegraphics{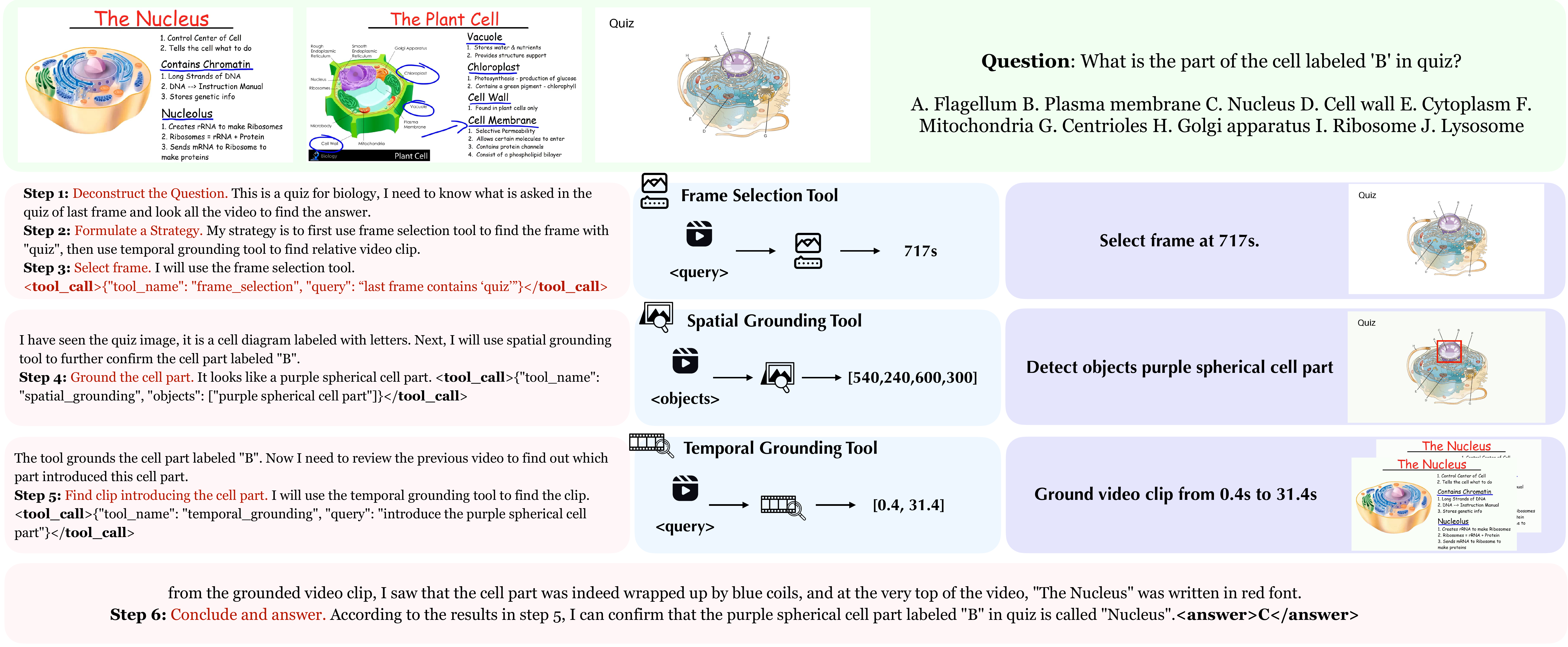}
  }
  % \vspace{-15pt}
  \caption{
    Visualization result of Weaver. The \textcolor{red}{\textbf{red}} regions indicate the model responses, the \textcolor{blue}{\textbf{blue}} regions denote the tool-calling processes, and the \textcolor{violet}{\textbf{purple}} regions correspond to the newly inserted visual information.
    }
  \label{fig: visualization}
  \vspace{-1.5em}
\end{figure*}

\subsection{In-Depth Analysis}
\noindent \textbf{Ablation on tool usage.}
\begin{table}[ht]
    % \vspace{-0.8em}
    \caption{Ablation results of different tool usage. TG: Temporal Grounding, FS: Frame Selection, TR: Trim, TC: Temporal Count, ST: Spatial Tracking, SG: Spatial Grounding.
    }
    \label{tab: tool ablation}
    \vspace{-5pt}
    \centering
    \renewcommand{\arraystretch}{1.1} % 调整行距
    \footnotesize
    % \small
    % \vspace{-1em}
    \setlength{\tabcolsep}{1.5mm}
    \begin{tabular}{lccccccc}
            \toprule
            \multicolumn{6}{c}{\textbf{Tools}} & \multirow{2}{*}{\textbf{LVReason}} & \multirow{2}{*}{\textbf{VideoMME}}\\
            TG & FS & TR & TC & ST & SG & & \\ \midrule
             & & & & & & 68.7 & 63.3\\
            \cmark & & & & & & 72.0 & 64.1\\
            \cmark & \cmark & & & & &  72.5 & 64.4\\
            \cmark & \cmark & \cmark & & & & 73.7 & 64.0\\
            \cmark & \cmark & \cmark & \cmark & & &  73.5 & 64.5\\
            \cmark & \cmark & \cmark & \cmark & \cmark & &  74.9& 65.1\\
            \cmark & \cmark & \cmark & \cmark & \cmark & \cmark & \textbf{75.4} & \textbf{65.3}\\
            \bottomrule
        \end{tabular}
        % \vspace{-2em}
\end{table}
To evaluate the effectiveness of each tool, we do an ablation to quantify them. We sequentially exclude training data containing a certain tool from the SFT dataset, and then training all variants in same epochs for RL stage. As demonstrated in Table~\ref{tab: tool ablation}, as the number of available tools gradually decreases, the overall performance of the system also shows a downward trend. Among them, the absence of the temporal tools has a greater impact on performance, indicating that temporal information is a crucial component in video reasoning tasks.
What's more, removing the spatial grounding tool alone does not significantly affect performance. However, when all tools related to spatial reasoning are eliminated, the system's performance drops relatively noticeably. This may be because when a single type of tool is removed, tools with similar functions can partially compensate for its absence.

\noindent \textbf{Ablation on agentic system.}
Considering prior work such as VideoAgent~\cite{fan2024videoagent} and ViperGPT~\cite{suris2023vipergpt}, it is feasible to directly construct a zero-shot agentic system using our tool library to solve problems. Accordingly, we conduct an ablation study to compare zero-shot agentic systems with our proposed method, thereby demonstrating the necessity of end-to-end training. 
As shown in Table~\ref{tab: agentic system ablation}, compared to the base model, the zero-shot agentic system also achieves improved performance on LVReason and VideoMME. Moreover, the effectiveness is positively correlated with the capability of the planner; the more powerful the planner, the higher the accuracy attained. Nonetheless, Weaver consistently outperforms all other agentic systems across these benchmarks, demonstrating the superiority of our approach.

\begin{table}[ht]
    \caption{Ablation results of zero-shot agentic system.
    }
    \label{tab: agentic system ablation}
    \vspace{-5pt}
    \centering
    \renewcommand{\arraystretch}{1.1} % 调整行距
    \footnotesize
    % \small
    % \vspace{-1em}
    \setlength{\tabcolsep}{1.1mm}
    \begin{tabular}{llccc}
            \toprule
            \textbf{Planner}& \textbf{Answerer}&\textbf{Trainable} & \textbf{LVReason} & \textbf{VideoMME}\\
            \midrule
            \xmark & Qwen2.5-VL & \xmark & 74.0 & 60.0\\ \hline
            Qwen2.5-VL & Qwen2.5-VL& \xmark & 76.0 & 64.0\\
            Gemini-2.5-Pro & Qwen2.5-VL & \xmark & 79.0 & 66.0\\
            Weaver & Weaver & \cmark & \textbf{81.0} & \textbf{67.0}\\
            \bottomrule
        \end{tabular}
        % \vspace{-2em}
\end{table}

\noindent \textbf{Ablation on training strategies.}
As demonstrated in Table~\ref{tab: strategy ablation}, we compare the pure text-centric Chain-of-Thought method with our proposed Weaver system. For a fair comparison, both methods are trained on the same data (Weaver-SFT and Weaver-RL); however, for the text-centric Chain-of-Thought model, we retrieve raw textual CoTs from the original dataset to serve as SFT supervision. The results in Table~\ref{tab: strategy ablation} indicate that Weaver consistently outperforms the text-centric Chain-of-Thought method, thereby underscoring the necessity of visual-text interleaved reasoning.

Furthermore, we conducted additional experiments to investigate training data strategies, focusing primarily on two aspects: the impact of various data sources and the importance of implementing a filtering strategy to obtain high-quality data. As shown in Table~\ref{tab: data ablation}, combining LVreason and VideoR1 datasets for both long and short videos leads to notable performance improvements compared with training on either dataset alone. Moreover, the absence of a filtering strategy results in a significant decline in performance.

\noindent \textbf{Analysis of tool usage.}
We also examine tool usage during the evaluation stage. As illustrated in Figure~\ref{fig:eval tool analysis}, we quantify both the average number of tools used per sample and the proportion of each tool employed across benchmarks. Figure~\ref{fig:sub1} demonstrates that, following the two-stage training, Weaver successfully learns to utilize multiple tools to solve problems, clearly indicating the flexibility of the method.

\begin{table}[ht]
    \caption{Ablation results of different training strategies.
    }
    \label{tab: strategy ablation}
    \vspace{-5pt}
    \centering
    \renewcommand{\arraystretch}{1.1} % 调整行距
    \footnotesize
    % \small
    % \vspace{-1em}
    \setlength{\tabcolsep}{2.2mm}
    \begin{tabular}{lcc}
            \toprule
            \textbf{Setting} & \textbf{LVReason} & \textbf{VideoMME}\\
            \midrule
            Baseline & 68.7 & 63.3\\ \hline
            Text-centric CoT & 74.3 & 60.0\\
            Weaver-SFT & 71.7 & 63.8\\
            Weaver & \textbf{75.4} & \textbf{65.3}\\
            \bottomrule
        \end{tabular}
        \vspace{-2em}
\end{table}

\begin{table}[ht]
    \caption{Ablation results of data strategies.
    }
    \label{tab: data ablation}
    \vspace{-5pt}
    \centering
    \renewcommand{\arraystretch}{1.1} % 调整行距
    \footnotesize
    % \small
    % \vspace{-1em}
    \setlength{\tabcolsep}{1.5mm}
    \begin{tabular}{lccccc}
            \toprule
            \multicolumn{2}{c}{\textbf{Data}} & \multirow{2}{*}{\textbf{Filter}} & \multirow{2}{*}{\textbf{LVReason}} & \multirow{2}{*}{\textbf{VideoMME}}\\
            LVR & VideoR1 & & & \\ \midrule
            \xmark & \xmark & \xmark & 68.7 & 63.3\\
            \cmark & \cmark & \xmark & 74.5 & 64.7\\
            \cmark & \xmark & \cmark & 74.4 & 65.0\\
            \xmark & \cmark & \cmark & 70.3 & 63.9\\
            \cmark & \cmark & \cmark & \textbf{75.4} & \textbf{65.3}\\
            \bottomrule
        \end{tabular}
        % \vspace{-2em}
\end{table}

Furthermore, Figure~\ref{fig:sub2} reveals that the system has learned to select different types of tools according to the specific requirements of distinct problems. For example, VSIBench is a benchmark that focuses on spatial reasoning; therefore, the frequency of frame selection and spatial grounding tool usage in this benchmark is significantly higher than in others, as it necessitates the analysis of static spatial relationships.

\vspace{-5pt}
\section{Conclusion}
\label{sec: conclusion}
This paper introduces Weaver, an end-to-end multimodal reasoning agentic system capable of employing a variety of tools to incrementally acquire visual evidence and generate interleaved visual-text reasoning trajectories. Our approach utilizes a two-stage training paradigm, empowering the system to autonomously discover optimal strategies for combining and utilizing these tools in a flexible manner. By mastering tool usage, the model gains access to progressive spatio-temporal visual information, thereby effectively improving performance. Extensive experiments on various complex video benchmarks demonstrate the system’s superior capabilities. We believe Weaver marks a significant step forward in the pursuit of Artificial General Intelligence.

{
    \small
    \bibliographystyle{ieeenat_fullname}
    \bibliography{main}
}
\onecolumn
{
    \centering
    \Large
    \textbf{Weaver: End-to-End Agentic System Training for Video Interleaved Reasoning}\\
    \vspace{0.5em}Appendix \\
    \vspace{1.0em}
}
\setcounter{page}{1}
\appendix

\section{Experimental Details}
\subsection{Training Details}
For the cold-start SFT training stage, we utilize Qwen’s official fine-tuning codebase~\cite{bai2025qwen2_5vl}. The learning rate is set to 1e-5, with a total batch size of 32 and a warm-up rate of 0.1. Training is conducted on 8 H800 GPUs over 2 epochs.

For the RL training stage, we modify the official codebase of verl~\cite{sheng2025hybridflow} to implement the multi-turn multimodal agentic reinforcement learning. 8 H800 GPUs are used for tool deployment and 8 H800 GPUs for model training. The detailed configuration is provided in Table~\ref{tab:supp_train_config}.

\begin{table*}[h]
    \centering
    \small
    \setlength{\tabcolsep}{6mm}
    \begin{tabular}{l|c}
        \toprule
        \textbf{Configuration} & \textbf{RL}\\
        \midrule
        method &  Tool-augmented GRPO \\
        freeze\_visual\_encoder & True \\
        learning\_rate & 1e-6 \\
        kl\_loss\_coef ($\beta$) & 1e-3 \\
        warmup\_ratio & 0 \\
        group\_size & 8 \\
        batch\_size & 64 \\
        mini\_batch\_size & 32 \\
        micro\_batch\_size\_per\_device & 1 \\
        max\_num\_turns & 10 \\
        max\_prompt\_length & 8192 \\
        max\_response\_length & 20480 \\
        \bottomrule
    \end{tabular}
    \caption{Training configurations.
    Group\_size is the number of rollouts, max\_num\_turns is the maximum number of conversation turns.
    }
    \label{tab:supp_train_config}
\end{table*}

\vspace{-1em}
\subsection{Agentic Tools Details}
In this section we introduce the details about the tools we use in the Weaver agentic system.

\vspace{3pt} \noindent \textbf{UniTime}~\cite{unitime2025}
UnTime is a SOTA video temporal grounding model built upon Qwen2-VL. During inference, UniTime performs iterative grounding based on the duration of the video, it can achieve precise temporal localization even for long videos by employing a step-by-step process that refines results from coarse to fine. We extracted the features of the required video in advance to improve efficiency.

\vspace{3pt} \noindent \textbf{Qwen2.5-VL}~\cite{bai2025qwen2_5vl}
We select Qwen2.5-VL-7B to serve as the expert model for two tasks within our agentic system: frame selection and temporal count.

For the frame selection tool, we employ Qwen2.5-VL to perform batch inference on the input frame sequence. Specifically, we instruct the model to provide a confidence score for each frame, indicating the degree to which it matches the input query. The frame with the highest confidence score is then selected as the final output.

For the temporal count tool, we divide the raw video into several clips and prompt the model to determine whether the queried event occurs within each clip. The final output is generated by stitching together clips in which the event is detected.

\vspace{3pt} \noindent \textbf{GroundedSAM2}~\cite{ren2024grounded}
GroundedSAM2 is a tracking model that integrates Grounding-Dino~\cite{liu2024grounding} and SAM2~\cite{ravi2024sam2}. The model first utilizes Grounding-Dino to convert the natural language query into bounding boxes within the starting frame. Subsequently, SAM2 is employed to track the objects contained within these bounding boxes throughout the sequence.

\vspace{3pt} \noindent \textbf{Grounding-Dino}~\cite{liu2024grounding}
Grounding-Dino is an open-vocabulary detection model that employs a Transformer-based architecture to integrate both text and image inputs, enabling the detection of objects based on free-form textual descriptions.

For the spatial grounding tool, we ground the objects in each frame individually when multiple frames are provided.

\section{Prompts}
In this section, we present the prompts use in our agentic system which contains the data construction prompt and inference system prompt.
\subsection{Prompt for Data Construction}
For Weaver-SFT-10K, we employ this prompt to invoke Gemini-2.5-Pro, enabling the rewriting of the text-centric Chain-of-Thought into a tool-augmented reasoning trajectory.
\begin{lstlisting}[language=Python]
You are an expert AI assistant specializing in multimodal reasoning and video analysis. Your primary function is to transform a raw thinking process into a structured, step-by-step reasoning chain. This chain must clearly articulate the analytical steps required to answer a question based on visual evidence from a video, including the strategic use of analytical tools.
Your Task
Given a question, a raw thinking process, and a list of available tools, you will generate a structured output in four distinct parts:
Question: The user\'s question, decoupled from the options.
Options: The multiple-choice options provided with the question.
Correct Answer: The ground truth answer.
Thinking Process with Tool Usage: A detailed, step-by-step explanation of the reasoning process.
"Thinking Process with Tool Usage" Requirements
This is the most critical part of your output. It must adhere to the following rules:
Format: It must be a list of JSON objects. Each object represents a single step in your thought process and must follow this exact format: {"from": "assistant", "content": "Description of the thinking step"}.
Visual Evidence Only: Your entire reasoning process must be based exclusively on the visual evidence that would be obtained by analyzing the video with the provided tools. Your logic should demonstrate how you infer the answer by analyzing the visual events in the video alone.
Strategic Tool Usage: For each step, you must critically assess whether a tool is genuinely needed.
Do not propose using a tool for steps that can be resolved through direct observation, logical deduction, or synthesis of information from previous steps.
When you do use a tool, you must explicitly state why it is necessary and what specific visual evidence it will help you gather.
Final Answer: The very last object in the JSON list must contain only the final answer, enclosed in <answer> tags (e.g., <answer>C</answer>).
EXAMPLE
[RAW_QUESTION]: Why does the man in white lift the child in blue up?  \nA. The child is crying and asking for help. \nB. The child looks up at the metal bar and wants to grasp it.  \nC. The man is helping the child get down from the bar.  \nD. The man is moving the child to a different play area.  \n
[THINK_PROCESS]: Step 1: Observe the child in blue\'s actions. At the beginning of the clip, the child in blue runs towards a horizontal metal bar. The child then jumps up, attempting to grab the bar but doesn\'t quite reach it.
Step 2: Observe the man\'s actions. A man in a white shirt approaches the child. As the child jumps and briefly hangs from the bar, the man reaches up and lifts the child higher, helping them get a better grip and position on the bar.
Step 3: Evaluate the given options based on the observations.
A. The child is crying and asking for help. There is no visual indication that the child is crying. Their actions (running and jumping) suggest excitement and a desire to play.
B. The child looks up at the metal bar and wants to grasp it. This aligns perfectly with the observed actions. The child runs directly to the bar and jumps to grab it, clearly showing their intent. The man\'s action is a direct response to this attempt, assisting them.
C. The man is helping the child get down from the bar. This is incorrect. The man is lifting the child up to the bar, not helping them get down.
D. The man is moving the child to a different play area. This is incorrect. The entire interaction is focused on the child playing on that specific horizontal bar.
Step 4: Conclude the best answer. Based on the analysis, the man lifts the child because the child showed a clear desire to grasp and play on the metal bar.
Therefore, the correct option is B.
[AVAILABLE_TOOLS]
Temporal grounding tool: Grounds a detailed temporal window of a certain event according to the "query". Usage: <tool_call>{"tool_name": "temporal_grounding", "query": "the event you want to ground."}</tool_call>
Spatial tracking tool: Tracks certain objects in spatial bounding boxes. Usage: <tool_call>{"tool_name": "spatial_tracking", "objects": ["object1", "object2", ...]}</tool_call>
Frame selection tool: Selects a single, most representative keyframe that best matches a textual query. This is ideal for answering questions about static scenes that do not require temporal analysis, such as counting objects or identifying attributes. Usage: <tool_call>{"tool_name": "frame_selection", "query": "a textual description of the desired scene or moment."}</tool_call>
Spatial grounding tool: Locates specified objects with bounding boxes in a more specific static scene. This tool operates on the current focused context, which should be a static scene (ideally a single frame) prepared by a preceding tool like frame_selection or a very short trim. If the context contains multiple frames, it will default to analyzing the middle frame. Usage: <tool_call>{"tool_name": "spatial_grounding", "objects": ["object1", "object2", ...]}</tool_call>
Trim tool: Grounds a detailed temporal window if you can get direct start and end timestamps from question, options or thinking process. It is also useful when you want to look at the video clip which is "before" or "after" some events, the default "start" is 0 and "end" is the duration of the video if you don\'t provide. The timestamp should be transfered into seconds. Usage: <tool_call>{"tool_name": "trim", "start": start timestamp, should be a float, "end": end timestamp, should be a float.}</tool_call>
Temporal Count tool: Ground a detailed event which can happen multiple times in the video, the tool will ground all related clip and concat them together. It should be used when the question explicitly requires count the number of some events. Usage: <tool_call>{"tool_name": "temporal_count", "query": "the event you want to count."}</tool_call>
You need to consider carefully which tool to use for similar type for different question type. (temporal grounding, temporal count and frame selection) (spatial tracking and spatial grounding)
(MODEL OUTPUT FOR THE EXAMPLE)
{
"Question": "Why does the man in white lift the child in blue up?",
"Options": {
"A": "The child is crying and asking for help."
"B": "The child looks up at the metal bar and wants to grasp it."
"C": "The man is helping the child get down from the bar."
"D": "The man is moving the child to a different play area."
},
"Correct Answer": "B",
"Thinking Process":
[
{
"from": "assistant",
"content": "Step 1: Analyze the Question. The question asks \'Why\' the man lifts the child, which requires me to infer his motivation. I must do this by establishing a cause-and-effect relationship between the child\'s actions and the man\'s subsequent action."
},
{
"from": "assistant",
"content": "Step 2: Formulate a Strategy. My strategy is to first use the available tools to determine the child\'s goal immediately before the man\'s intervention. Then, I will analyze the man\'s action to see if it helps the child achieve that specific goal."
},
{
"from": "assistant",
"content": "Step 3: Gather Evidence of Child\'s Intent. I will use the temporal grounding tool to isolate the child\'s actions just before the man helps. \n\n<tool_call>{\n  \"tool_name\": \"temporal_grounding\",\n  \"query\": \"the child in blue running towards and jumping to reach the yellow horizontal bar\"\n}</tool_call>\n\nThis tool would return the video segment where the child looks up at the bar, runs toward it, and jumps with outstretched arms. This is strong visual evidence that the child\'s objective is to grasp the bar."
},
{
"from": "assistant",
"content": "Step 4: Gather Evidence of Man\'s Action. Next, I will use the temporal grounding tool to analyze the man\'s interaction with the child. \n\n<tool_call>{\n  \"tool_name\": \"temporal_grounding\",\n  \"query\": \"the man in white lifting the child in blue up to the yellow bar\"\n}</tool_call>\n\nThe tool would show the man lifting the child vertically, directly towards the same yellow bar the child was trying to reach. This action directly addresses the child\'s apparent goal."
},
{
"from": "assistant",
"content": "Step 5: Synthesize and Evaluate Options. The visual evidence shows a clear sequence: the child tries and fails to reach the bar, and then the man lifts the child to the bar. Based on this, I will evaluate the options."
},
{
"from": "assistant",
"content": "Step 6: Evaluate Option A. The grounded clips of the child\'s face show no visual signs of crying or distress. Therefore, this option is incorrect."
},
{
"from": "assistant",
"content": "Step 7: Evaluate Option B. This option states the child wants to grasp the bar. The child\'s upward gaze, run, and jump are all visual cues that support this intent. The man\'s action directly helps achieve this. This option is correct."
},
{
"from": "assistant",
"content": "Step 8: Evaluate Option C. The man\'s motion is upward, bringing the child to the bar, not downward from the bar. This option contradicts the visual evidence."
},
{
"from": "assistant",
"content": "Step 9: Evaluate Option D. A spatial tracking tool on the man and child would show they remain in the same location by the play structure. The man is not moving the child to a different area. This option is incorrect."
},
{
"from": "assistant",
"content": "<answer>B</answer>"
}]}
YOUR TASK
[RAW_QUESTION]: <raw_questions>
[THINK_PROCESS]: <cot>
[AVAILABLE_TOOLS]
Temporal grounding tool: Grounds a detailed temporal window of a certain event according to the "query". Usage: <tool_call>{"tool_name": "temporal_grounding", "query": "the event you want to ground."}</tool_call>
Spatial tracking tool: Tracks certain objects in spatial bounding boxes. Usage: <tool_call>{"tool_name": "spatial_tracking", "objects": ["object1", "object2", ...]}</tool_call>
Frame selection tool: Selects a single, most representative keyframe that best matches a textual query. This is ideal for answering questions about static scenes that do not require temporal analysis, such as counting objects or identifying attributes. Usage: <tool_call>{"tool_name": "frame_selection", "query": "a textual description of the desired scene or moment."}</tool_call>
Spatial grounding tool: Locates specified objects with bounding boxes in a more specific static scene. This tool operates on the current focused context, which should be a static scene (ideally a single frame) prepared by a preceding tool like frame_selection or a very short trim. If the context contains multiple frames, it will default to analyzing the middle frame. Usage: <tool_call>{"tool_name": "spatial_grounding", "objects": ["object1", "object2", ...]}</tool_call>
Trim tool: Grounds a detailed temporal window if you can get direct start and end timestamps from question, options or thinking process. It is also useful when you want to look at the video clip which is "before" or "after" some events, the default "start" is 0 and "end" is the duration of the video if you don\'t provide. The timestamp should be transfered into seconds. Usage: <tool_call>{"tool_name": "trim", "start": start timestamp, should be a float, "end": end timestamp, should be a float.}</tool_call>
Temporal Count tool: Ground a detailed event which can happen multiple times in the video, the tool will ground all related clip and concat them together. It should be used when the question explicitly requires count the number of some events. Usage: <tool_call>{"tool_name": "temporal_count", "query": "the event you want to count."}</tool_call>
Now, generate the new question, options, correct answer, and the detailed "Thinking Process with Tool Usage" for the new video provided above. Remember to base the raw thinking process and the visual evidence from the tools.
\end{lstlisting}

\subsection{System Prompt}
\label{sec: system prompt}
To enhance Weaver’s understanding of its assigned tasks, we design a system prompt that introduces each tool and clarifies the objective of invoking these tools to answer the questions.

\begin{lstlisting}[language=Python]
You are a helpful multimodal assistant. Your task is to solve complex visual questions by thinking step-by-step and using tools.
#Tools
You are provided with following tools:
1. Temporal grounding tool: Grounds a detailed temporal window of a certain event according to the "query". Usage: <tool_call>{"tool_name": "temporal_grounding", "query": "the event you want to ground."}</tool_call>
2. Spatial tracking tool: Tracks certain objects in spatial bounding boxes. Usage: <tool_call>{"tool_name": "spatial_tracking", "objects": ["object1", "object2", ...]}</tool_call>
3. Frame selection tool: Selects a single, most representative keyframe that best matches a textual query. This is ideal for answering questions about static scenes that do not require temporal analysis, such as counting objects or identifying attributes. Usage: <tool_call>{"tool_name": "frame_selection", "query": "a textual description of the desired scene or moment."}</tool_call>
4. Spatial grounding tool: Locates specified objects with bounding boxes in a more specific static scene. This tool operates on the current focused context, which should be a static scene (ideally a single frame) prepared by a preceding tool like frame_selection or a very short trim. If the context contains multiple frames, it will default to analyzing the middle frame. Usage: <tool_call>{"tool_name": "spatial_grounding", "objects": ["object1", "object2", ...]}</tool_call>
5. Trim tool: Grounds a detailed temporal window if you can get direct start and end timestamps from question, options or thinking process. It is also useful when you want to look at the video clip which is "before" or "after" some events, the default "start" is 0 and "end" is the duration of the video if you don't provide. The timestamp should be transfered into seconds. Usage: <tool_call>{"tool_name": "trim", "start": start timestamp, should be a float, "end": end timestamp, should be a float.}</tool_call>
6. Temporal Count tool: Ground a detailed event which can happen multiple times in the video, the tool will ground all related clip and concat them together. It should be used when the question explicitly requires count the number of some events. Usage: <tool_call>{"tool_name": "temporal_count", "query": "the event you want to count."}</tool_call>
#Instructions
1. In each step, you need to give a decomposed thinking process, and evaluate whether it is needed to use tools and which tools to use.
2. You need to consider carefully which tool to use for similar type for different question type. (temporal grounding, temporal count and frame selection) (spatial tracking and spatial grounding)
3. After calling tools and getting return results, you need to analyze the results and judge whether the results is useful. If not, you can recall the tool with different parameters.
4. The results obtained from the tool may not always be accurate. You need to carefully watch the newly obtained fragments and analyze whether there is content you need. If not, proceed to the next step of analysis.
5. If you think the process is ended and no more steps in needed, you need to output the final answer in <answer></answer> tags. e.g. <answer>Answer here</answer>.
\end{lstlisting}

% \vspace{-1em}
\section{More Results}
In this section, we provide additional visualizations to illustrate the reasoning process of our Weaver agentic system. These examples further demonstrate the superiority of our approach in solving various types of problems. By flexibly integrating the use of different tools, Weaver is able to progressively acquire accurate visual information and make correct judgments.

% \vspace{-1em}
\begin{figure*}[h]
  \resizebox{\linewidth}{!}{
  \includegraphics{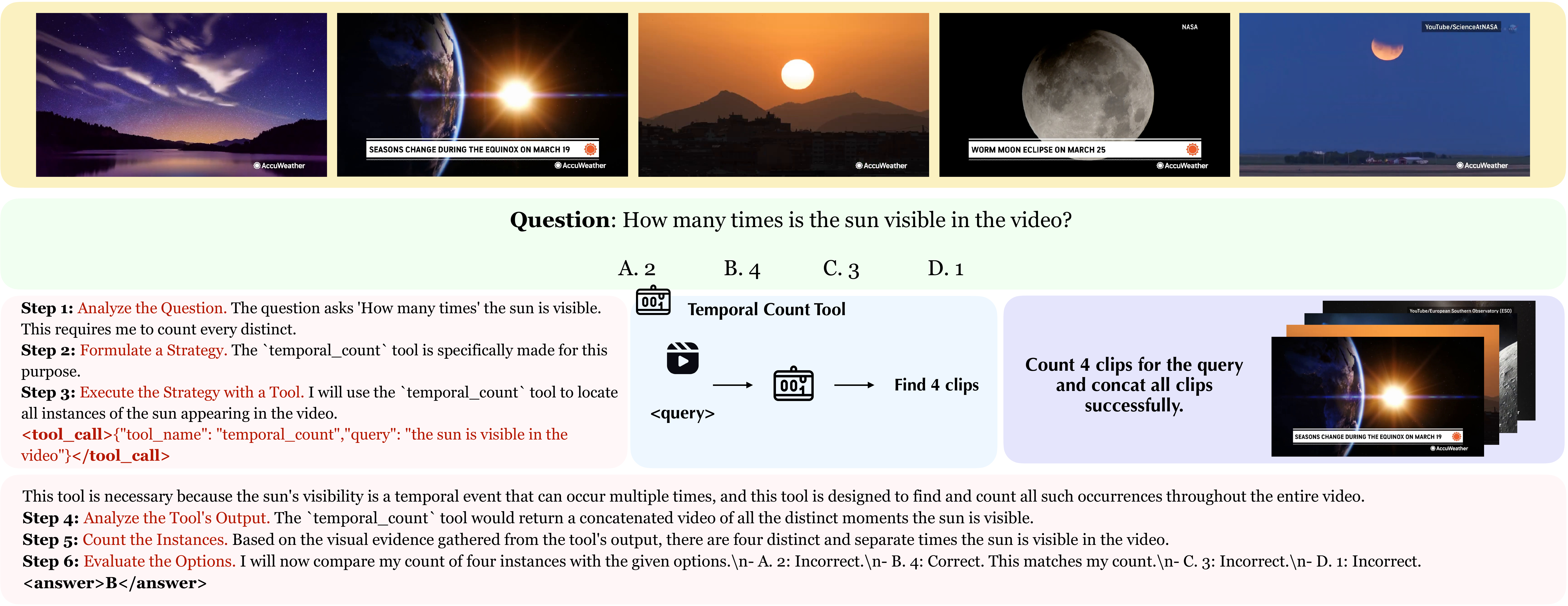}
  }
  \caption{
    Example from VideoMME~\cite{fu2025videomme}.
  }
  % \vspace{-1em}
\end{figure*}

\begin{figure*}[h]
  \resizebox{\linewidth}{!}{
  \includegraphics{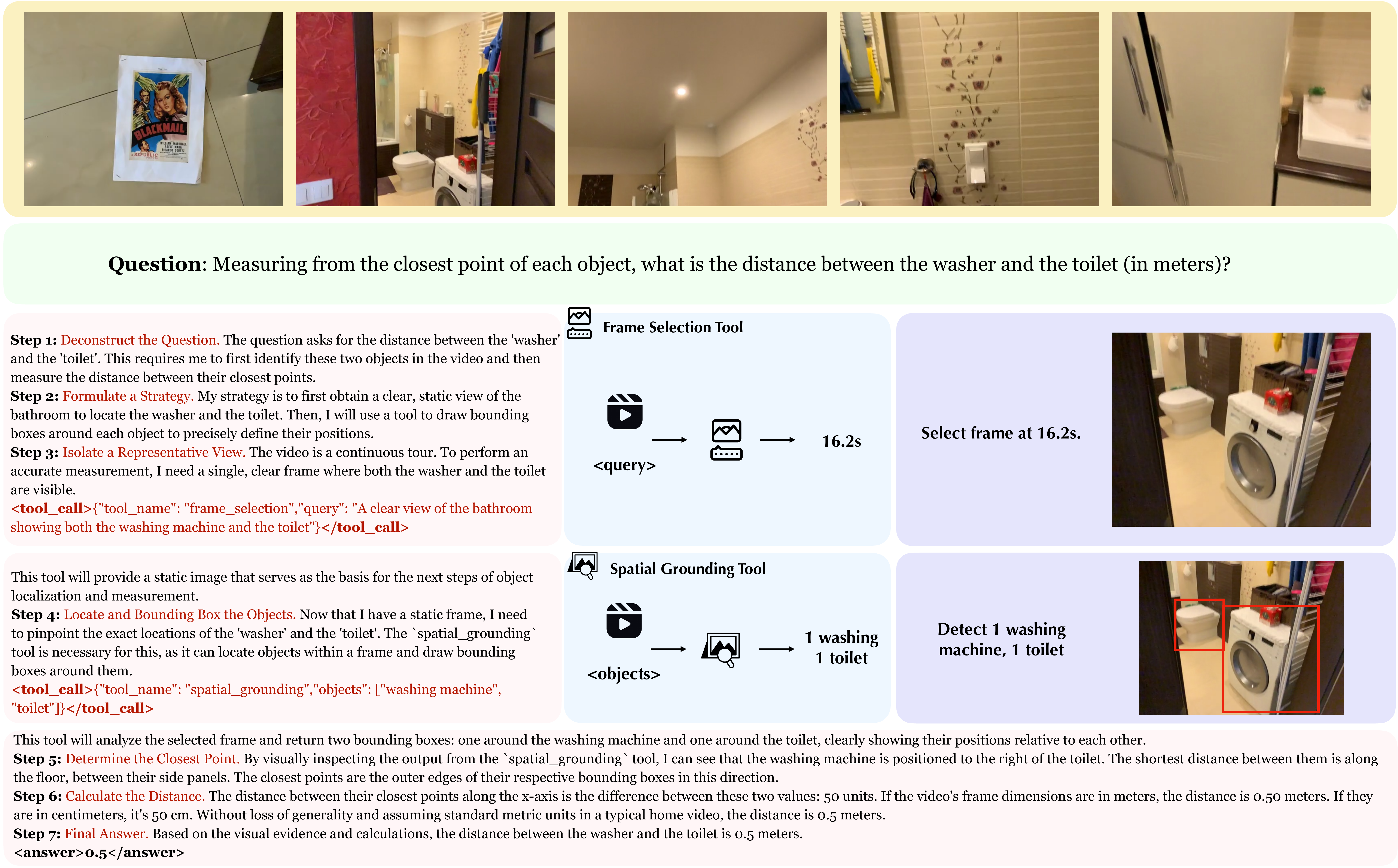}
  }
  \caption{
    Example from VSIBench~\cite{yang2025thinking}.
  }
  % \vspace{-1.1em}
\end{figure*}

\begin{figure*}[h]
  \resizebox{\linewidth}{!}{
  \includegraphics{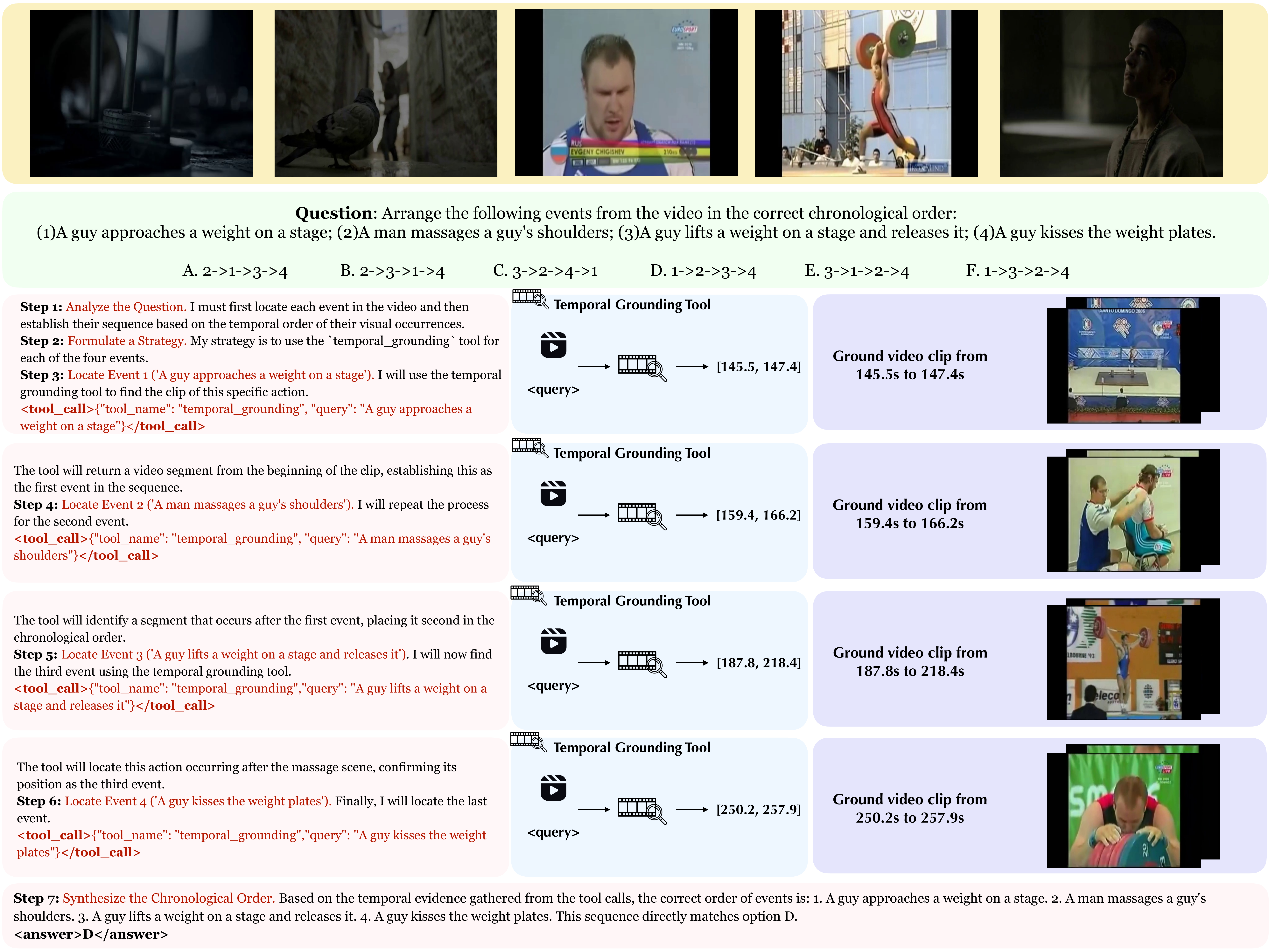}
  }
  % \vspace{-5pt}
  \caption{
    Example from MLVU~\cite{zhou2025mlvu}.
  }
  % \vspace{-1.1em}
\end{figure*}

\begin{figure*}[h]
  \resizebox{\linewidth}{!}{
  \includegraphics{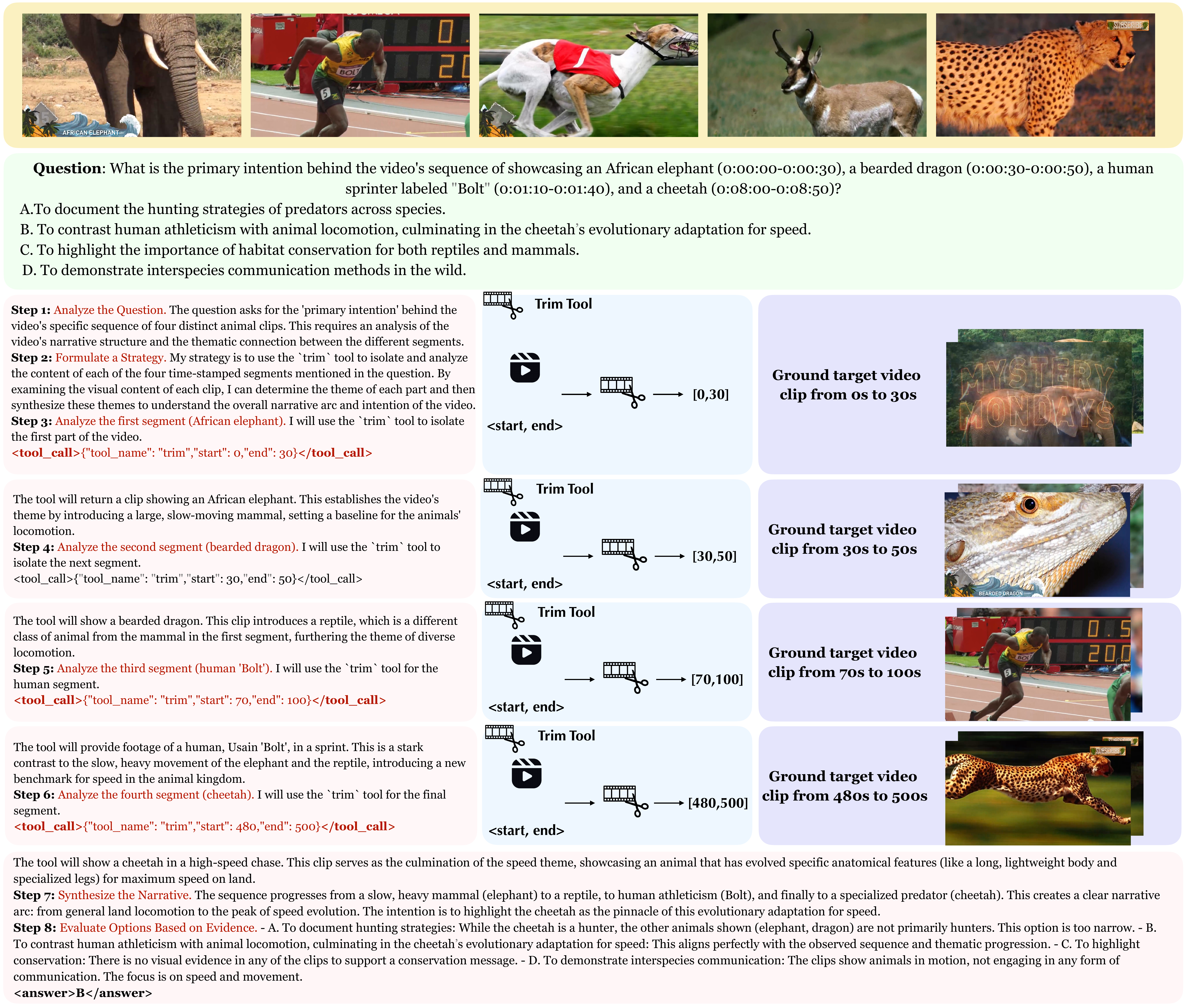}
  }
  % \vspace{-5pt}
  \caption{
    Example from LVReason~\cite{chen2025longrl}.
  }
  % \vspace{-1.1em}
\end{figure*}

% WARNING: do not forget to delete the supplementary pages from your submission 
% \input{sec/X_suppl}

\end{document}